# The 3D Structural Phenotype of the Glaucomatous Optic Nerve Head and its Relationship with The Severity of Visual Field Damage

Fabian A. Braeu[1,2,3], Thanadet Chuangsuwanich[1,3], Tin A. Tun[4,5], Shamira A. Perera[4,5], Rahat Husain[4], Aiste Kadziauskiene[6,7], Leopold Schmetterer[4,5,8-12], Alexandre H. Thiéry[13], George Barbastathis[2,14], Tin Aung[3,4,5], and Michaël J.A. Girard[1,5,12]

1. Ophthalmic Engineering & Innovation Laboratory, Singapore Eye Research Institute, Singapore National Eye Centre, Singapore
2. Singapore-MIT Alliance for Research and Technology, Singapore
3. Yong Loo Lin School of Medicine, National University of Singapore, Singapore
4. Singapore Eye Research Institute, Singapore National Eye Centre, Singapore
5. Duke-NUS Graduate Medical School, Singapore
6. Clinic of Ears, Nose, Throat and Eye Diseases, Institute of Clinical Medicine, Faculty of Medicine, Vilnius University, Vilnius, Lithuania
7. Center of Eye Diseases, Vilnius University Hospital Santaros Klinikos, Vilnius, Lithuania
8. SERI-NTU Advanced Ocular Engineering (STANCE), Singapore, Singapore
9. School of Chemistry, Chemical Engineering and Biotechnology, Nanyang Technological University Singapore
10. Department of Clinical Pharmacology, Medical University of Vienna, Austria
11. Center for Medical Physics and Biomedical Engineering, Medical University of Vienna, Austria
12. Institute of Molecular and Clinical Ophthalmology, Basel, Switzerland
13. Department of Statistics and Applied Probability, National University of Singapore, Singapore
14. Department of Mechanical Engineering, Massachusetts Institute of Technology, Cambridge, Massachusetts 02139, USA

**Corresponding Author:** Michaël J.A. Girard
Ophthalmic Engineering & Innovation Laboratory (OEIL)
Singapore Eye Research Institute (SERI)
The Academia, 20 College Road
Discovery Tower Level 6,
Singapore 169856
mgirard@ophthalmic.engineering
https://www.ophthalmic.engineering

# Abstract

**Purpose:** To describe the 3D structural changes in both connective and neural tissues of the optic nerve head (ONH) that occur concurrently at different stages of glaucoma using traditional and AI-driven approaches.

**Design:** Retrospective cross-sectional study.

**Methods:** We included 213 normal, 204 mild glaucoma (mean deviation [MD] ≥ -6.00 dB), 118 moderate glaucoma (MD of -6.01 to -12.00 dB), and 118 advanced glaucoma patients (MD < -12.00 dB). All subjects had their ONHs imaged in 3D with Spectralis optical coherence tomography. To describe the 3D structural phenotype of glaucoma as a function of severity, we used two different approaches: **(1)** We extracted 'human-defined' 3D structural parameters of the ONH (total of 10) including retinal nerve fiber layer (RNFL) thickness, minimum rim width, lamina cribrosa (LC) shape and depth at different stages of glaucoma; **(2)** we also employed a geometric deep learning method (i.e. PointNet) to identify the most important 3D structural features that differentiate ONHs from different glaucoma severity groups without any human input.

**Results:** We observed that the majority of ONH structural changes occurred in the early glaucoma stage, followed by a plateau effect in the later stages. Using PointNet, we also found that 3D ONH structural changes were present in both neural and connective tissues. Specifically, 57% (normal to mild glaucoma), 39% (mild to moderate glaucoma), and 53% (moderate to advanced glaucoma) of ONH landmarks that showed major structural changes were located in neural tissues with the remaining located in connective tissues. In both approaches, we observed that structural changes were more prominent in the superior and inferior quadrant of the ONH, particularly in the RNFL, the prelamina, and the LC. As the severity of glaucoma increased, these changes became more diffuse (i.e. widespread), particularly in the LC.

**Conclusions:** In this study, we were able to uncover complex 3D structural changes of the ONH in both neural and connective tissues as a function of glaucoma severity. We hope to provide new insights into the complex pathophysiology of glaucoma that might help clinicians in their daily clinical care.

# Introduction

Evaluation of structural changes of the optic nerve head (ONH) – the main site of damage in glaucoma – is a crucial step in diagnosing and monitoring glaucoma [1, 2]. The complex three-dimensional (3D) morphological changes occurring in glaucomatous ONHs can be captured and quantified by optical coherence tomography (OCT) – a fast, high-resolution, quantitative, and non-invasive 3D imaging modality [3].

In current medical practice, several investigations are conducted to assess neural tissue health. These tests involve both a functional (e.g., visual field testing) and a structural assessment of glaucomatous damage. The latter is typically achieved by measuring the thickness of the retinal nerve fiber layer (RNFL) via OCT [4-6]. Researchers have further investigated the association between other neural structural parameters with glaucomatous visual field damage, such as the thickness of the ganglion cell complex (GCC) [7, 8] and Bruch's membrane opening - minimum rim width (BMO-MRW) [9, 10].

However, recent research has indicated that the pathophysiology of glaucoma is multifaceted and cannot purely be characterized as damage to retinal ganglion cells: (1) Brooks et al. reported that the characteristic "glaucomatous cupping" of the ONH cannot solely be explained by neural tissue loss [11]; (2) Quigley et al. found that glaucomatous changes of the lamina cribrosa (LC) precede visual field damage [12]; and (3) Yang et al. suggested that ONH connective tissue deformations are the primary cause of retinal ganglion cell axonal injury [13]. These studies indicate that the pathophysiology of glaucoma should consider the involvement of the biomechanics, mechanobiology, remodeling, and potential mechanical breakdown of ONH connective tissues. Given this new understanding, researchers have begun to investigate the association between ONH connective tissue changes and glaucoma severity through connective tissue parameters extracted from OCT images of the ONH. Examples of such parameters include the LC depth (LCD) and the LC global shape index (LC-GSI) [14], the thickness of the peripapillary choroid [15], the scleral canal opening [16], and the peripapillary scleral angle representing the amount of bowing of the ONH [17]. However, no study has yet provided a comprehensive analysis of 3D structural changes of both the connective and neural tissues of the ONH that occur concurrently at different stages of glaucoma.

Therefore, the aim of this study was to describe the 3D structural phenotype of glaucoma as a function of severity by: **(1)** Extracting neural and connective tissue ONH parameters from segmented 3D OCT scans and investigating their differences between glaucoma severity groups; **(2)** Using 3D point clouds representing the complex structure of the ONH as an input for a geometric deep learning technique (i.e. PointNet [18]) that allows us to identify the major 3D structural changes of the ONH with glaucoma severity. Overall, we hope that our work leads to a better understanding of the pathophysiology of glaucoma that might improve the diagnosis and prognosis of glaucoma.

## Methods

### Patient Recruitment

This retrospective study involved a total of 414 subjects with glaucoma and 213 controls without glaucoma from two different cohorts: **(1)** 541 subjects of Chinese ethnicity were recruited at the Singapore National Eye Centre (SNEC) as part of their standard clinical care and **(2)** 112 subjects of European descent were recruited at the Vilnius University Hospital Santaros Klinikos as part of a prospective observational study. All subjects gave written informed consent and the study adhered to the tenets of the Declaration of Helsinki and was approved by the institutional review board of the respective institutions (SingHealth Centralized Institutional review board, Singapore and Vilnius Regional Biomedical Research Ethics Committee, Lithuania).

**Standard Automated Perimetry**

All subjects had their visual field (VF) assessed by standard automated perimetry (SAP; Swedish Interactive Threshold Algorithm standard 24-2 or 30-2 program; Humphrey Field Analyzer II-750i, Carl Zeiss Meditec). Subjects with a non-reliable VF examination that was defined using the criteria of a false-positive error rate greater than 15% [19] and a fixation loss greater than 33% [19, 20] were excluded from the study.

**Definition of Glaucoma and Glaucoma Severity Groups**

Glaucomatous eyes were defined as those with vertical cup-disc ratio (VCDR) > 0.7 and/or neuroretinal rim narrowing with repeatable glaucomatous VF defects and non-occludable angles on gonioscopy whereas non-glaucomatous (normal) eyes were those with an IOP < 21 mmHg and normal VF examinations. Subjects with corneal abnormalities that potentially can reduce the quality of the OCT scans and with ONH disorders other than glaucoma were excluded from the studies.

Based upon the mean deviation (MD) of the 24-2 or 30-2 VF, all glaucoma subjects were further split into three glaucoma severity groups [21]: **(1)** mild glaucoma (MD $\geq$ -6.00 dB); **(2)** moderate glaucoma (MD of -6.01 to -12.00 dB); and **(3)** advanced glaucoma (MD $<$ -12.00 dB). Even though this classification has its limitations [22], it remains a standard and can be used as a good first indicator for staging functional damage. More information on the demographics of the four groups (i.e. normal, mild, moderate, and advanced) can be found in **Table 1**.

**Optical Coherence Tomography Imaging**

Each patient from both cohorts had their ONH imaged with the same spectral domain OCT device (Spectralis, Heidelberg Engineering, Germany). All OCT scans (horizontal raster scans) covered an area of 15°x10° centered on the ONH and the number of B-scans varied between 49 and 97 B-scans (distance between B-scans varied from approximately 35 to 70 µm) with 384 A-scans per B-scan (approximately 11.5 µm between A-scans) and 496 pixels per A-scan (axial resolution of 3.87 µm/pixel). Images were acquired using signal averaging, eye tracking, and the enhanced depth imaging modality of the Spectralis OCT device.

**Describing the Structural Phenotype of Glaucoma as a Function of Glaucoma Severity**

In the following sections, we introduce two different approaches to study the complex structural changes of the ONH as a function of glaucoma severity. In the first, we performed a comprehensive 3D structural analysis of the ONH using 'human-defined' 3D structural parameters of the ONH (total of 10 parameters) describing the morphologies of both neural and connective tissues. In the second, we used a relatively recent geometric deep learning method (i.e. PointNet) to discover important 3D structural features differentiating ONHs from different glaucoma severity groups. An overview of both approaches is shown in **Figure 1**.

**Approach 1 for Describing the Structural Phenotype of Glaucoma – ONH Parameters**

For this approach, all ONH tissues were segmented in 3D (from the OCT scans), from which all ONH structural parameters were automatically extracted.

**AI-based Segmentation of ONH Tissues.** We automatically segmented all raw OCT volume scans of the ONH using REFLECTIVITY (Reflectivity, Abyss Processing Pte Ltd, Singapore) – a software that was developed from advances in AI-based ONH segmentation [23] (see **Figure 1a, b**). More specifically, we automatically labelled the following ONH tissue groups: **(1)** the retinal nerve fiber layer (RNFL) and the prelamina tissue (PLT); **(2)** the ganglion cell inner plexiform layer (GCL+IPL); **(3)** all other retinal layers (ORL); **(4)** the retinal pigment epithelium (RPE) with Bruch's membrane (BM) and the BM opening (BMO) points; **(5)** the choroid; **(6)** the OCT-visible part of the peripapillary sclera including the scleral flange; and **(7)** the OCT-visible part of the LC. In almost all OCT volume scans, the posterior boundaries of the sclera and LC were not visible and could therefore not be segmented.

**Automated extraction of ONH parameters.** Using the software REFLECTIVITY, we extracted the following parameters: **(1)** the average RNFL thickness (RNFLT) in each octant (i.e. temporal [T], superior-temporal [ST], superior [S], superior-nasal [SN], nasal [N], inferior-nasal [IN], inferior [I], and inferior-temporal [IT]) calculated at a distance of 1.5 times the BMO radius (BMOR) from the centre of BMO; **(2)** the average minimum rim width (MRW) in each octant defined as the minimum distance from a BMO point to a point on the inner limiting membrane (ILM); **(3)** the average ganglion cell complex thickness (GCCT) in each octant evaluated at the same location than the RNFLT; **(4)** the average choroidal thickness (ChT) in each octant at the same distance as that used for the RNFLT; **(5)** the prelamina depth (PLD) defined as the distance from the BMO center to a point on the ILM (perpendicular to the BMO plane); **(6)** the minimum prelamina thickness (MPT); **(7)** the LC depth (LCD) defined as the distance from the BMO centre to a point on the anterior LC boundary (perpendicular to the BMO plane); **(8)** the LC global shape index (LC-GSI) that summarizes the shape of the anterior LC boundary into a single number [24]; **(9)** the peripapillary scleral angle (PPSA) representing the amount of scleral bowing and defined as the angle between two parallel lines to the anterior scleral boundary in the nasal-temporal plane; and **(10)** the BMO area defined as the area of the best-fit ellipse to the BMO points. A visualization of the extracted ONH parameters is shown in **Figure 1c**.

**Statistical analysis.** All parameters were compared across all 4 groups (normal, mild, moderate, and advanced). All statistical analyses were performed using R (version 4.2.1) and RStudio (version 2022.07.1 for macOS). ONH parameters that were extracted in each octant were reported as mean $\pm$ standard deviation and single valued ONH parameters were presented as box plots. One-way ANOVA with post-hoc Tukey HSD test was used for the comparisons. P value for significance was set at <0.05.

**Approach 2 for Describing the Structural Phenotype of Glaucoma – PointNet**

PointNet, a deep neural network from the group of geometric deep learning algorithms, can learn from complex 3D shapes, such as that of the ONH, if they are represented as 3D point clouds. In contrast to our first approach, which relied on 'human-defined' ONH parameters, PointNet allows us to identify important structural landmarks that can differentiate ONHs from the four different glaucoma severity groups, without previous inputs or guidance.

**Representation of the ONH structure as 3D point cloud.** We described the structure of a given ONH as a 3D point cloud which then was used as input to PointNet. To do so, we first identified the anterior boundaries of all tissue layers in the segmented OCT scan. Each anterior boundary voxel was then represented as a 3D point (see **Figure 1d**). The final point cloud consisted of about 20,000 points for each ONH (see **Figure 1e**). Additionally, for each point, we extracted the local tissue thickness (minimum distance between anterior and posterior boundary). In summary, we assigned four values to every point: its position in the 3D space ([x, y, z]-coordinate) and its local tissue thickness (not applicable for the sclera and LC). To homogenize the data across all ONHs, the centre of BMO was set as origin of the coordinate system [x=0, y=0, z=0] and the normal of BMO plane (best-fit plane to the BMO points) was aligned with the axial direction of the scan. The more interested reader is referred to our previous publication on geometric deep learning for glaucoma diagnosis [25].

**Glaucoma severity classification.** PointNet was specifically designed to process and learn from 3D point clouds such as the one shown in **Figure 1**. We used the same architecture as in the original publication [18], except that we implemented a max pooling layer of dimension 256. To identify important 3D structural features of the ONH at different stages of glaucoma, we trained three PointNet classification networks to differentiate between: **(1)** normal and mild glaucoma subjects (normal-mild); **(2)** mild and moderate glaucoma subjects (mild-moderate); and **(3)** moderate and advanced glaucoma subjects (moderate-advanced).

To assess the performance of the three binary classification networks, we split each respective dataset (i.e. normal-mild, mild-moderate, and moderate-advanced) in training (70%), validation (15%), and test (15%) sets. To improve performance and reduce overfitting, we used data augmentation techniques such as random cropping, random rotations, random rigid translations, random sampling (i.e. randomly picking a subset of points from the input point cloud), oversampling to reduce data imbalance, and additive Gaussian noise where applicable. A five-fold cross validation study was performed (using the train and validation set) to tune hyperparameters and we reported the area under the receiver operating characteristic curves (AUCs) of the model with the best performing hyperparameters as mean ± standard deviation. All models were trained on a Nvidia RTX A5000 GPU card until optimum performance was reached in the validation set.

**Identification of important 3D structural features of the ONH.** The specific architecture of PointNet inherently allowed us to identify regions of the ONH important for the differentiation of different glaucoma severity groups by extracting all points that contributed to the final classification score – the so-called critical points. For each classification group (i.e. normal-mild, mild-moderate, and moderate-advanced), we extracted critical points from all ONHs of the respective test set (networks trained on respective training set using tuned hyperparameters). Comparing the locations of these points between the three groups allowed us to draw conclusion on the characteristic 3D structural changes of the ONH at different stages of glaucoma.

**Visualization of critical points.** To better visualize the location of the resulting critical points, we first constructed an average ONH geometry (represented by the average anterior boundaries of each segmented tissue) for each of the three classification groups, i.e. normal-mild, mild-moderate, and moderate-advanced. For each group, we then projected the critical points (closest point projection) onto their corresponding anterior tissue boundary of the respective average ONH geometry and visualized them as 3D point cloud density maps. A density measure for each point was obtained by counting the neighbouring points within a 75 µm radius sphere. Since all critical points were projected on an average ONH geometry, such a density map should highlight landmarks of the ONH that exhibit distinct 3D structural changes between the different stages of glaucoma (represented as a cluster of red points in the point cloud density maps).

# Results

## Approach 1 – Statistical Analysis of ONH Parameters

We observed that the majority of ONH structural changes occurred in the early glaucoma stage (normal to mild). These changes were also the most substantial in terms of their size or magnitude. Specifically, we noted a decrease in average RNFLT (average over all sectors) from 112 ± 26 µm to 83 ± 29 µm (**Figure 2a**), a decrease in average MRW from 256 ± 60 µm to 169 ± 55 µm (**Figure 2b**), a decrease in average GCCT from 154 ± 26 µm to 124 ± 30 µm (**Figure 2c**), no change in average ChT (**Figure 2d**), an increase in PLD from 136 ± 195 µm to 288 ± 199 µm (**Figure 2e**), a decrease in MPT from 146 ± 116 µm to 63 ± 70 µm (**Figure 2f**), an increase in LCD from 410 ± 109 µm to 468 ± 132 µm (**Figure 2g**), a decrease in LC-GSI from -0.37 ± 0.42 to -0.61 ± 0.33 (**Figure 2h**), an increase in PPSA from 5.4 ± 4.6 degree to 9.5 ± 6.2 degree (**Figure 2i**), and an increase in BMOA from 2.15 ± 0.5 mm$^2$ to 2.28 ± 0.5 mm$^2$ (**Figure 2j**).

Following substantial structural changes of the ONH in the early stage of glaucoma, most ONH parameters showed a plateau effect, with little change from mild to moderate glaucoma. Only RNFLT (average), GCCT (average), and MRW (average) showed a significant decrease from 83 ± 29 to 71 ± 30 µm, 124 ± 30 to 111 ± 32 µm, and 169 ± 55 to 159 ± 56 µm, respectively.

In the later stages of glaucoma (moderate to advanced), we observed significant structural changes of the ONH, but they were much less pronounced in terms of their magnitude compared to those seen in the early stages. In detail, the average RNFLT decreased from 71 ± 30 µm to 50 ± 25 µm (**Figure 2a**), the average MRW decreased from 159 ± 56 µm to 126 ± 46 µm (**Figure 2b**), the average GCCT decreased from 111 ± 32 µm to 88 ± 27 µm (**Figure 2c**), the LCD increased from 459 ± 121 to 502 ± 147 µm (**Figure 2g**), and the BMOA decreased from 2.30 ± 0.58 mm$^2$ to 2.12 ± 0.42 mm$^2$ (**Figure 2j**). The ChT (**Figure 2d**), the PLD (**Figure 2e**), the MPT (**Figure 2f**), the LC-GSI (**Figure 2h**), and the PPSA (**Figure 2i**) showed no significant change.

If we were to examine regional variations, we noted that structural changes of the RNFLT, MRW, and GCCT were more pronounced (higher in magnitude) in both the superior and inferior octants of the ONH. This was true throughout all stages of glaucoma. In these sectors, we observed that the decrease in MRW slowed as glaucoma severity increased. Specifically, in the early stage of glaucoma (normal to mild), MRW decreased in the superior octant from 295 ± 64 µm to 192 ± 58 µm while in the later stage (moderate to advanced), the decrease was smaller from 179 ± 58 µm to 133 ± 49 µm (**Figure 2b**). In contrast, RNFLT and GCCT decreased linearly as glaucoma severity increased. In the early stage of glaucoma (normal to mild), RNFLT and GCCT in the superior octant decreased from 163 ± 31 to 122 ± 34 µm and 200 ± 31 to 160 ± 34 µm, respectively, while in the later stage (moderate to advanced), the decrease was from 102 ± 35 to 61 ± 31 µm and 141 ± 35 to 99 ± 32 µm (**Figure 2a, 2c**). With the exception of the inferior octant of the ONH, we did not observe any significant changes in the ChT with glaucoma severity (**Figure 2d**).

**Approach 2 – Performance Assessment**

Using PointNet, we were able to differentiate ONHs from different glaucoma severity groups. The normal-mild glaucoma classification showed the best performance (AUC: 0.94 ± 0.02), followed by the moderate-advanced (AUC: 0.80 ± 0.04) and mild-moderate glaucoma classification (AUC: 0.68 ± 0.08).

## Approach 2 – Changes of Important 3D Structural Features of the ONH with Glaucoma Severity

For each classification task (i.e. normal-mild, mild-moderate, and moderate-advanced), we pooled all critical points from all ONHs (test set), mapped them onto the corresponding average ONH geometry, and displayed them as a 3D point cloud density map for all ONH tissues (**Figure 3**), or separately for each ONH tissue (**Figure 4**).

In general, we observed that critical points were present in both, neural (normal-mild: 57%, mild-moderate: 39%, moderate-advanced: 53%) and connective tissues (normal-mild: 43%, mild-moderate: 61%, moderate-advanced: 47%). More specifically, most of the critical points were located in the RNFL+PLT (normal-mild: 53%, mild-moderate: 37%, moderate-advanced: 47%), the sclera (normal-mild: 17%, mild-moderate: 15%, moderate-advanced: 11%), and the LC (normal-mild: 23%, mild-moderate: 43%, moderate-advanced: 31%). In contrast, we observed almost no critical points in the other tissue layers, such as the GCC+IPL, ORL, RPE, Choroid.

On a tissue level, we found that the critical points from the RNFL of all three classification tasks formed an hourglass pattern with points mainly located in the superior and inferior quadrant. In addition, in the normal-mild glaucoma classification, critical points from the RNFL were mostly located around the neuro-retinal rim whereas in the moderate-advanced glaucoma classification, these points moved more outwards to the peripheral region of the ONH. Interestingly, we also found that in the normal-mild and mild-moderate classification most of the critical points from the LC were located near the LC insertion zone in the superior (normal-mild) and superior + inferior quadrant (mild-moderate) whereas in the moderate-advanced classification, critical points were more spread out over the entire LC.

## Discussion

In this study, we were able to describe the 3D structural phenotype of glaucoma as a function of severity using two separate approaches. In the first, we extracted 'human-defined' 3D structural parameters of the ONH and compared them across four different groups: normal, mild, moderate, and advanced. In the second, we represented the complex structure of the ONH as a 3D point cloud and used PointNet to uncover the structural landmarks that were the most affected by glaucoma severity without any human input. Overall, we found that the structural features of both neural and connective tissues contributed to the structural phenotype of glaucoma, and that each of our proposed method could provide its own unique knowledge.

In this study, we found that after substantial structural changes of the ONH in the early stage of glaucoma (normal to mild), almost all ONH parameters reached a plateau, with less change in the later stages (mild to moderate and moderate to advanced). This is in good agreement with previous studies that investigated the structure-function relationship and reported a considerable structural loss before any functional VF defects were detectable [26-28]. Some of these studies suggested a "tipping point" in the early stage of glaucoma (at about – 3 dB MD) from which onwards even small structural changes were associated with a relatively large decrease in MD value [26, 28]. One should also keep in mind that MD values are usually reported on a logarithmic scale (non-linear scale). For instance, a shift in MD value from 0 to -6 dB would imply a much larger loss in visual sensitivity compared to a shift from - 6 to -12 dB on a linear scale [29]. Therefore, the observed plateau effect might be a result of reporting MD values on a logarithmic scale. However, further research is needed to verify such a hypothesis.

Furthermore, we found that critical points were present in both neural (normal-mild: 57%, mild-moderate: 39%, moderate-advanced: 53%) and connective tissues (normal-mild: 43%, mild-moderate: 61%, moderate-advanced: 47%) at all stages of glaucoma indicating that the structural changes caused by glaucoma affected both types of tissue in the ONH. Our findings are in line with previous research that suggested that the pathophysiology of glaucoma is complex and cannot purely be characterized as a damage to the neural tissue in the ONH (i.e. retinal ganglion cells) [11-13]. Despite these recent findings, current glaucoma tests focus on assessing neural tissue health, ignoring any glaucomatous structural changes of connective tissue in the ONH. In the future, the development of more comprehensive tests that consider structural changes in both, neural and connective tissues, could potentially improve the diagnosis and prognosis of glaucoma.

Additionally, we found that most of the critical points (normal-mild: 93%, mild-moderate: 95%, moderate-advanced: 89%) were concentrated in the RNFL+PLT, sclera, and LC. PointNet only focuses on the major structural changes of the optic nerve head, and since we limited the number of critical points to 256, only the ONH landmarks with significant 3D structural changes will be highlighted in the point cloud density maps. Therefore, the fact that there are almost no critical points present in the GCC+IPL, ORL, RPE, and choroid does not necessarily imply that these tissues do not exhibit any structural changes in glaucoma. However, our findings suggest that any structural changes in these tissues are likely to be smaller in magnitude compared to the structural changes observed in the RNFL, sclera, and LC.

In both approaches, we found that structural changes of neural tissues were more prominent in the inferior and superior quadrants of the ONH over all stages of glaucoma. This is in accordance with many previous studies (including our recent study in glaucoma diagnosis [25]) that reported significant structural changes of glaucomatous ONHs in these quadrants [30, 31]. In addition, Wang et al. reported a progressive nasalization of the central retinal vessel trunk (CRVT) with glaucoma severity [32]. One might argue that the location of some of the critical points from the RNFL coincides with the location of the CRVT and its branches indicating changes in the CRVT location with disease progression. However, further research is needed to confirm such speculations.

Furthermore, we found that the decline in MRW slowed, whereas RNFLT decreased linearly as glaucoma severity increased. This suggests that neural tissue changes in the early stage of glaucoma (normal to mild) are more pronounced around the optic disc (i.e. MRW), in contrast to the later stages of glaucoma (mild to moderate and moderate to advanced), where such changes move to the periphery of the ONH (i.e. RNFLT). Interestingly, we found a similar trend in the distribution of critical points from the RNFL. In the early glaucoma group (normal-mild), critical points were mostly located around the neuro-retinal rim. These critical points (with their local tissue thickness) might act as a surrogate measurement for MRW. In the more severe glaucoma groups (i.e. mild-moderate and moderate-advanced), critical points from the RNFL moved to more peripheral regions of the ONH and thus closer to where the RNFLT was measured. Up to date, there is no common consent on whether RNFLT or MRW is better correlated with VF damage (i.e. glaucoma severity). Some studies favored RNFLT [10, 33] whereas others reported better performance of MRW [30, 34]. In addition, Gmeiner et al. reported that depending on the stage of glaucoma and the major site of glaucomatous damage (peripheral or central), RNFLT might be superior to MRW and vice versa suggesting that morphological changes of the glaucomatous ONH are diverse and may depend on various factors [33]. Therefore, when assessing ONH structural changes, it might be important to analyze the entire region of the ONH (peripheral and central) with its complex 3D morphology as it was done with PointNet.

We found that a considerable number of critical points were extracted from the sclera over all stages of glaucoma, suggesting significant and progressive structural changes of the sclera with glaucoma severity. In addition, and in line with a previous study [17], we found that the PPSA, representative for the bending of the sclera in the nasal-temporal plane, is significantly larger in mild glaucoma compared to normal eyes, however, no significant differences were found between the later stages of the disease. Considering the presence of critical points from the sclera in all stages of glaucoma, one might speculate that a single parameter like the PPSA is not enough to capture the complex 3D structural changes of the sclera with glaucoma severity and further research is needed to quantify such changes.

Furthermore, we found that most of the LC critical points were located in the region of the LC insertion zone over all stages of glaucoma. However, the major site of these critical points changed from the superior quadrant (normal-mild) to the superior + inferior quadrant (mild-moderate) to a more diffuse distribution over all quadrants (moderate-advanced). Previous studies reported morphological changes of the LC with glaucoma severity reflected by a change in LC depth [35], LC curvature [36], and LC-GSI [14]. In addition, local LC defects or alterations like posterior movement of the LC insertion zones [37] and LC disinsertions [38] were observed in glaucomatous eyes. However, none of the studies reported structural changes of the LC insertion zone with glaucoma severity. Our findings suggest that assessing morphological changes of the glaucomatous LC, especially in the region of the LC insertion zone, could be useful in monitoring disease progression (in conjunction with other ONH parameters like the RNFLT). However, further longitudinal studies are necessary to unravel the complex 3D structural changes of the LC with glaucoma severity.

In this study, several limitations warrant further discussion. First, although the overall sample size was fairly large, however, subjects were unevenly distributed over the glaucoma severity groups (normal: 213, mild: 204, moderate: 118, advanced: 118). In addition, the Caucasian subgroup had no healthy controls that might introduce a bias in both, the comparison of ONH parameters and the learning process of PointNet. Therefore, our findings might not be easily transferable to other populations. In the future, we want to investigate possible differences in structural changes of the ONH with glaucoma severity between different ethnic groups.

Second, we used MD values of the 24-2 or 30-2 VF to determine glaucoma severity, however, standard automated perimetry is subjective and sometimes underestimate disease severity [22]. Recent studies suggest chromatic pupillometry [39] or electroretinogram [40] as an objective way to assess functional loss in glaucomatous eyes. However, these devices have their own limitations and a future study has to show whether our findings would change when using a different staging system.

Third, the accuracy of the extracted ONH parameters and the extracted point clouds to represent local structural features of the ONH depends on the performance of the segmentation algorithm. Even though the segmentation software that we used in this study (Reflectivity, Abyss Processing Pte Ltd, Singapore) was tested and validated on a large cohort of glaucomatous and non-glaucomatous ONHs at different stages of glaucoma, one should keep in mind that the choice of the segmentation algorithm might have an impact on the results.

Fourth, although we found that many ONH parameters showed significant differences between glaucoma severity groups, the cross-sectional nature of our data limits causal inferences. As a result, our findings might differ from longitudinal studies that follow individual patients over a certain period of time. In the future, we aim to validate our findings by applying our herein developed approaches to a longitudinal dataset.

Fifth, the differentiation of ONHs from the mild and moderate glaucoma severity group was the most challenging task and resulted in a rather small AUC of $0.68 \pm 0.08$ (PointNet). The moderate performance of PointNet might be due to the plateau effect that we observed after substantial structural changes in the early stage of glaucoma. In the future, we could consider the MD value as a continuous variable and predict its "true" value, instead of a binary classification, as this might give us a boost in performance.

In summary, we successfully described the 3D structural phenotype of glaucoma as a function of glaucoma severity by: **(1)** a "traditional" approach based on extracted ONH parameters and **(2)** a more recently introduced approach based on critical points extracted by PointNet. We showed that ONH structural changes are not limited to neural tissues but occurred in both, neural and connective tissues simultaneously. In addition, we identified a major site of 3D morphological change of the ONH that might potentially be worth monitoring in the future - the region around the LC insertion zone. With this study, we hope to provide new insights into the complex pathophysiology of glaucoma that might help clinicians in their daily clinical care.

## Acknowledgment

We acknowledge funding from **(1)** the donors of the National Glaucoma Research, a program of the BrightFocus Foundation, for support of this research (G2021010S [MJAG]); **(2)** SingHealth Duke-NUS Academic Medicine Research Grant (SRDUKAMR21A6 [MJAG]); **(3)** the "Retinal Analytics through Machine learning aiding Physics (RAMP)" project that is supported by the National Research Foundation, Prime Minister's Office, Singapore under its Intra-Create Thematic Grant "Intersection Of Engineering And Health" - NRF2019-THE002-0006 awarded to the Singapore MIT Alliance for Research and Technology (SMART) Centre [MJAG/AT/GB]. **(4)** the "Tackling & Reducing Glaucoma Blindness with Emerging Technologies (TARGET)" project that is supported by the National Medical Research Council (NMRC), Singapore (MOH-OFLCG21jun-0003 [MJAG]).

# Figures

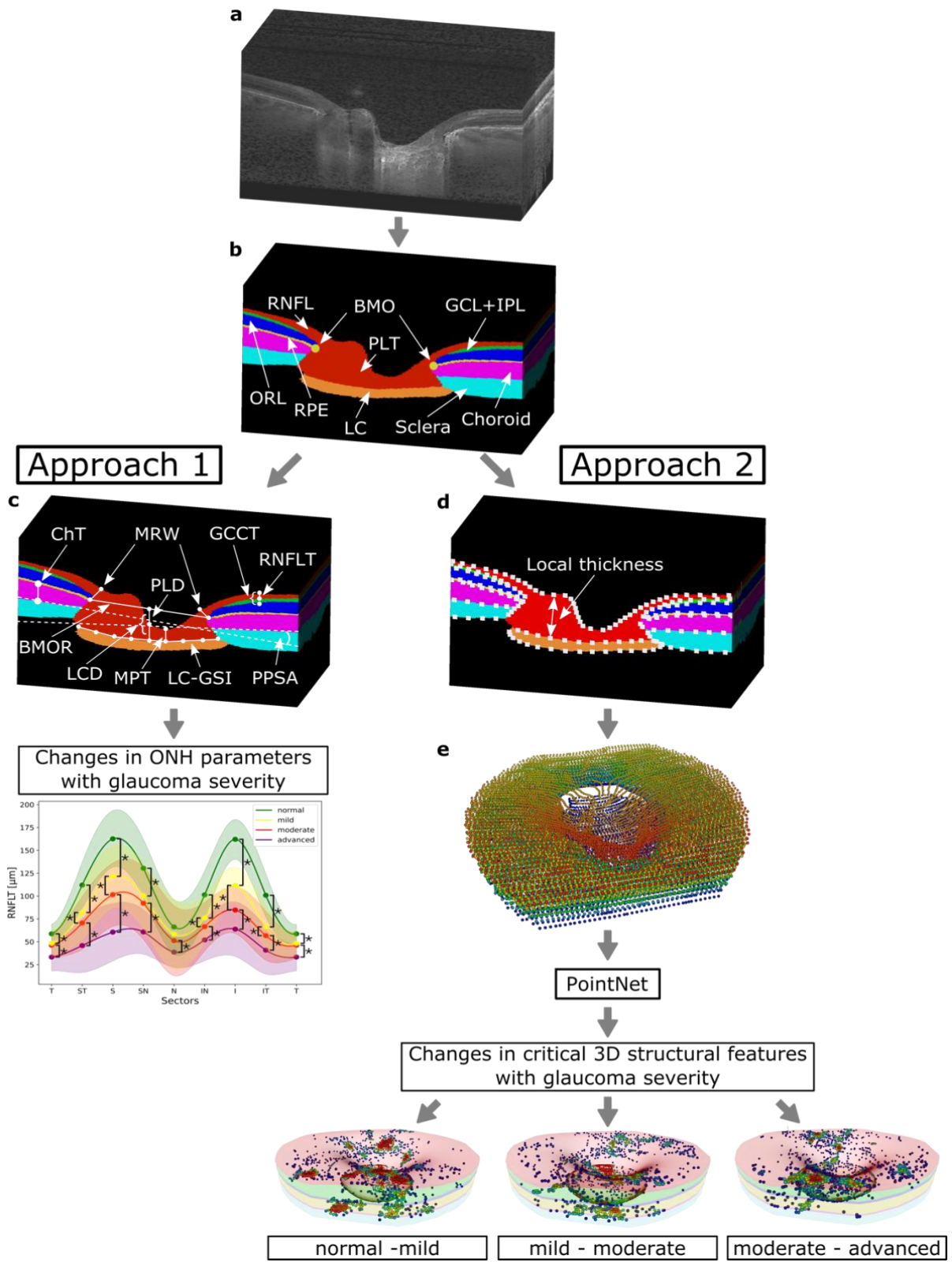

**Figure 1.** Overview of two approaches to describe the 3D structural phenotype of glaucoma as a function of severity. Approach 1 was based on the comparison of well-established ONH parameters between different glaucoma severity groups (**a**-**c**). Approach 2 leverages on geometric deep learning to identify important 3D landmarks of the ONH to differentiate ONHs at different stages of glaucoma. By looking at the changes of these critical 3D structural features with glaucoma severity, we were able to draw conclusions about the complex 3D structural changes of the ONH taking place at different stages of glaucoma (**a**, **b**, **d**, and **e**).

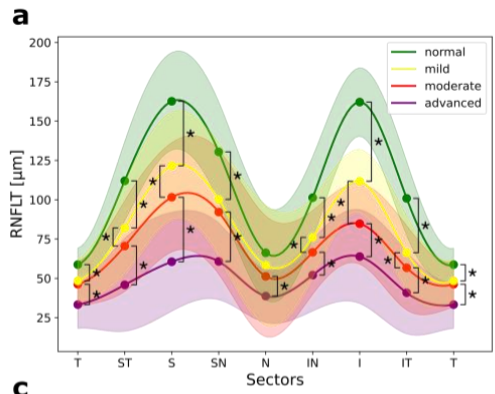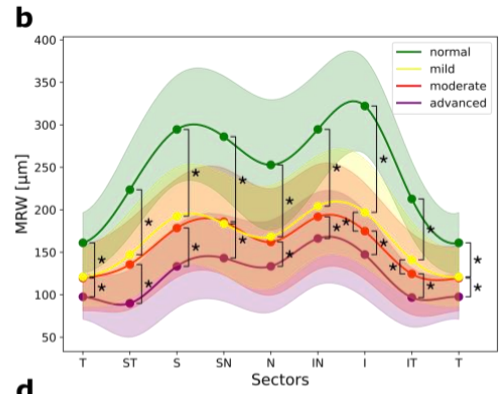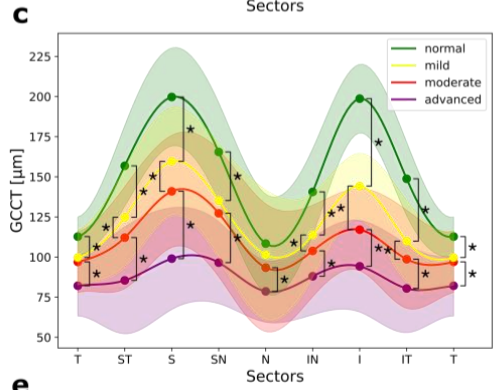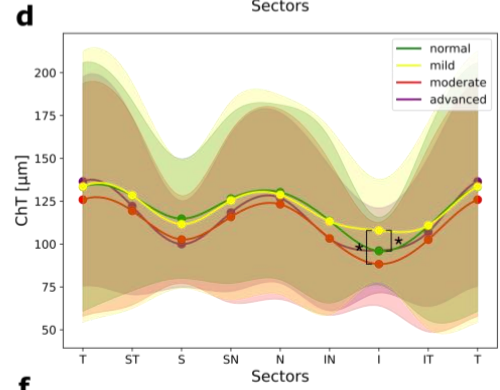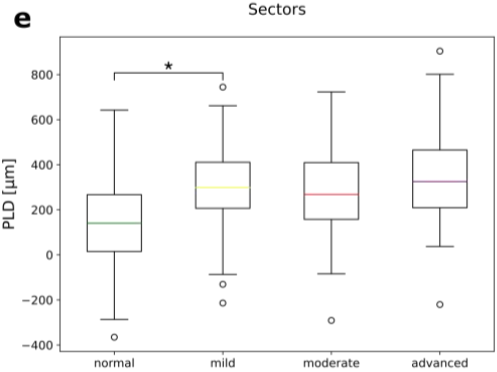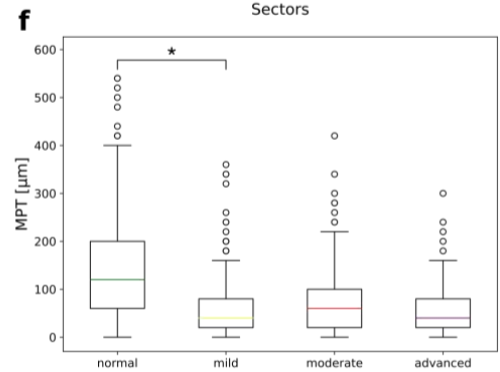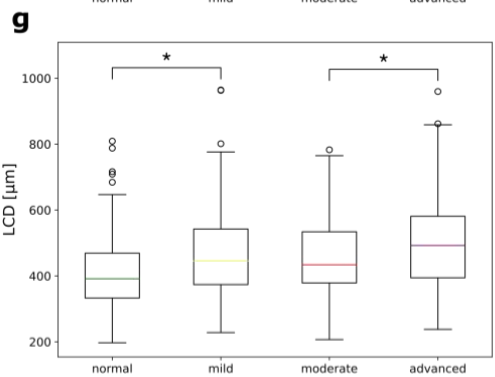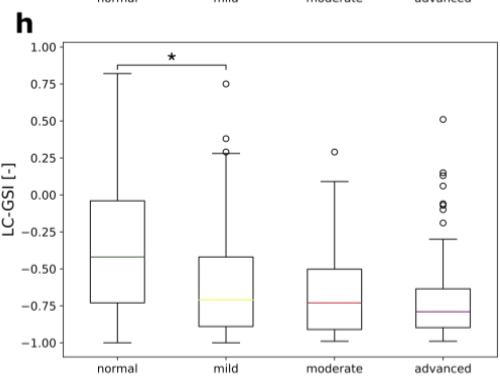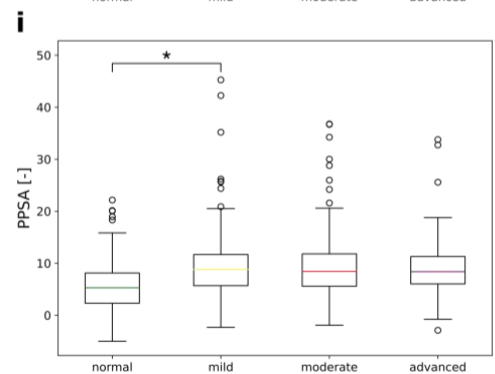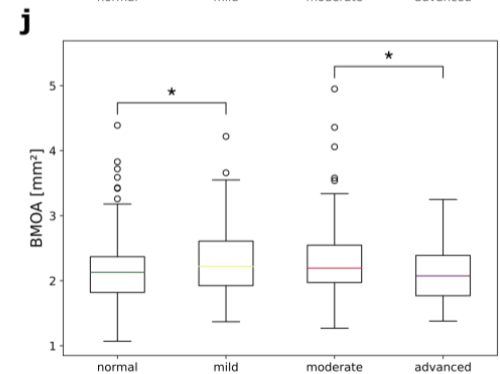

**Figure 2**. Summary of statistical analysis of automatically extracted ONH parameters. RNFLT, MRW, GCCT, and ChT are shown as sector plots (T: temporal, ST: superior-temporal, S: superior, SN: superior-nasal, N: nasal, NI: nasal-inferior, and I: inferior sector) with values for each group given as average ± standard deviation. Non-sectorial parameters are presented as boxplots. A significant difference between two groups (p<0.05) was indicated with a * (determined by post-hoc Tukey HSD tests).

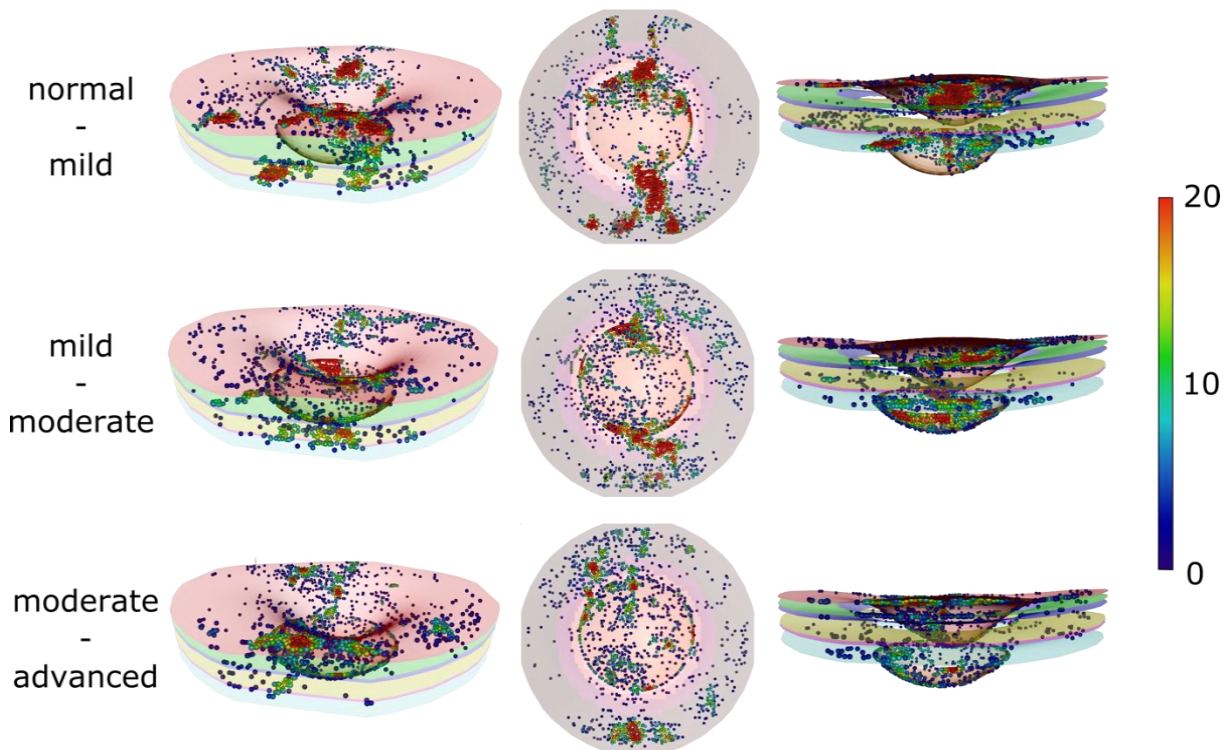

**Figure 3**. Critical points resulting from the three classification tasks: normal-mild, mild-moderate, and moderate advanced. From left to right column: 3D, en face (top), and sagittal (side) view. Surfaces represent the average anterior tissue boundaries for each respective dataset: RNFL+PLT (red), GCL+IPL (green), ORL (blue), RPE (yellow), choroid (purple), sclera (cyan), and LC (orange). Red colored critical points correspond to ONH regions with high importance for the differentiation of the respective glaucoma severity groups.

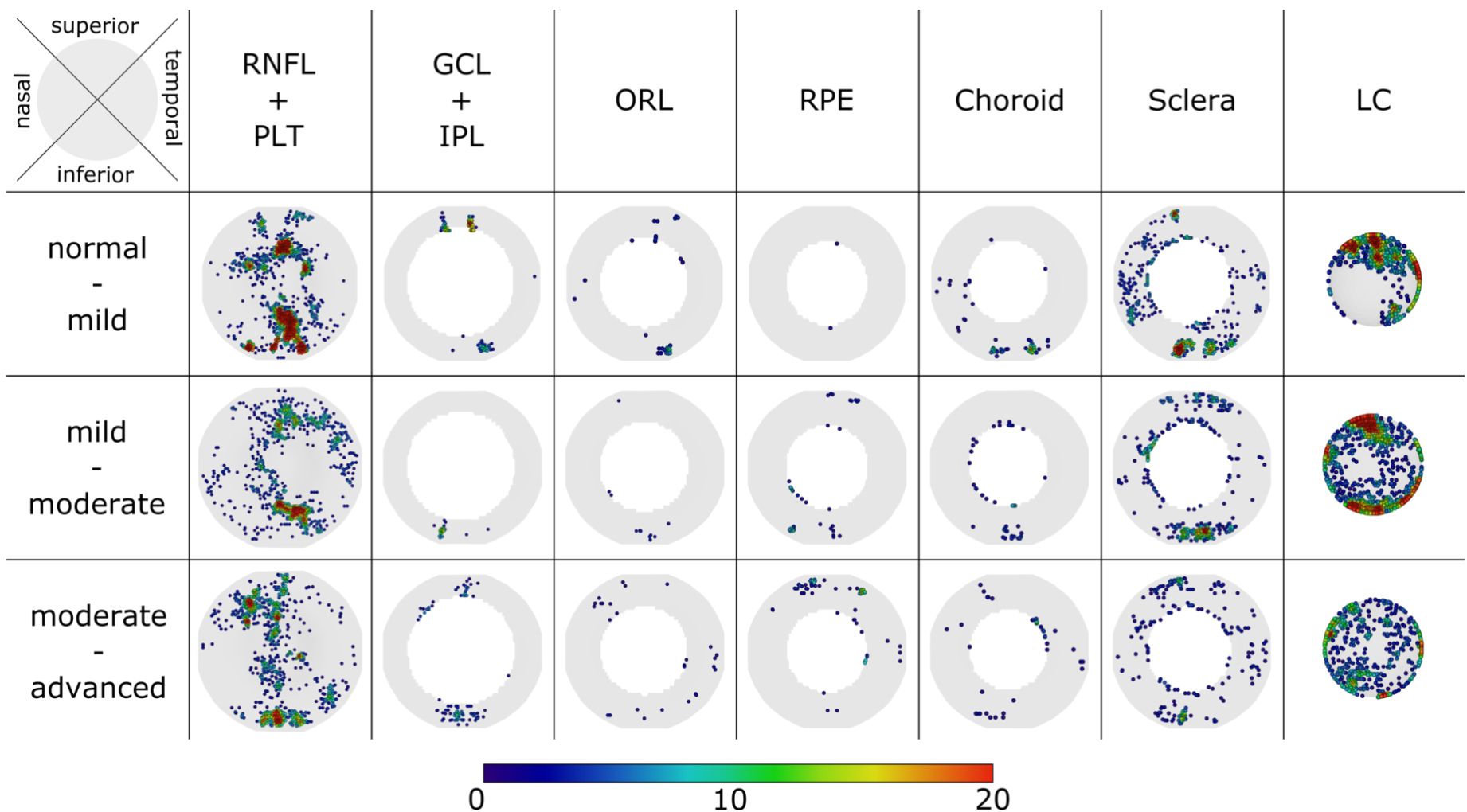

**Figure 4.** En face (top) view layer by layer comparison (columns) of critical points at different stages of glaucoma severity (rows). Critical points are presented as point cloud density maps with colours indicating the number of neighbouring points within a sphere with a radius of 75 µm.

5 # Tables

6
7 **Table 1.** Summary of glaucoma severity groups.
8

|  | NORMAL (N=213) | MILD (N=204) | MODERATE (N=118) | ADVANCED (N=118) | P* |
|---|---|---|---|---|---|
| **AGE, YEARS** | 63.36 (6.99) | 66.9 (6.42) | 68.05 (7.11) | 68.52 (7.69) | <0.001 |
| **SEX, FEMALE** | 126 (59.15) | 91 (44.61) | 49 (41.52) | 43 (36.44) | <0.001 |
| **RACE** |  |  |  |  |  |
|   CHINESE | 213 | 178 | 97 | 53 | <0.001 |
|   CAUCASIAN | 0 | 26 | 21 | 65 |  |
| **MD, DB** | -1.41 (2.11) | -3.35 (1.95) | -8.16 (2.35) | -18.64 (5.31) | <0.001 |

9 Data are in mean (standard deviation) or n (%) as appropriate.
10 MD = mean deviation of the 24-2 or 30-2 visual field test.
11 *Comparison between the four groups using Fisher's exact test (for sex and race) and ANOVA
12 (for age and MD).